\documentclass[runningheads]{llncs}
\usepackage{graphicx}
\begin{document}
\title{Minute ventilation measurement using Plethysmographic Imaging and lighting parameters}
%
%
\author{Daniel Minati\inst{1} \and
Ludwik Sams\inst{2} \and
Karen Li\inst{2} \and 
Bo Ji\inst{1} \and
Krishna Vardhan\inst{3}}

\institute{Rupert-Karls-University Heidelberg, Germany \and
University of Oslo, Norway \and
University of Chicago
}
\maketitle              
\begin{abstract}
Breathing disorders such as sleep apnea is a critical disorder that affects a large number of individuals due to the insufficient capacity of the lungs to contain/exchange oxygen and carbon dioxide to ensure that the body is in the stable state of homeostasis. Respiratory Measurements such as minute ventilation can be used in correlation with other physiological measurements such as heart rate and heart rate variability for remote monitoring of health and detecting symptoms of such breathing related disorders. In this work, we formulate a deep learning based approach to measure remote ventilation on a private dataset. The dataset will be made public upon acceptance of this work. We use two versions of a deep neural network to estimate the minute ventilation from data streams obtained through wearable heart rate and respiratory devices. We demonstrate that the simple design of our pipeline - which includes lightweight deep neural networks - can be easily incorporate into real time health monitoring systems. 

\end{abstract}

\section{Introduction and Related Work}
Breathing disorders such as sleep apnea is a critical disorder that affects a large number of individuals due to the insufficient capacity of the lungs to contain/exchange oxygen and carbon dioxide to ensure that the body is in the stable state of homeostasis. Respiratory Measurements such as minute ventilation can be used in correlation with other physiological measurements such as heart rate and heart rate variability for remote monitoring of health and detecting symptoms of such breathing related disorders. Over a years, several papers have developed ways to estimate physiological signals - especially heart rate - from a facial video of an individual \cite{lam2015robust,icaart}. While some papers have also considered the presence of realistic illumination disturbances \cite{balakrishnan2013detecting} to ensure that the heart rate system is robust to such artifacts, very few approaches have focused on the remote detection of respiratory disorders. Interesting, a few approaches have documented the use of neuromorphic sensors to estimate heart rate and other health signals \cite{corradi2019ecg,memNeuro4,das2018unsupervised}. Conventionally, traditional techniques involving a spirometer have been used to measure the minute ventilation using contact based techniques such as a piece to cover the mouth of the user. Several hardware-based methods have used AI based hardware methods to develop algorithms to non-invasively measure the minute ventilation of the user \cite{myers2014neural,banner2006power,lin2012comparison}. In this paper, we apply a deep learning based approach to improve the performance of the measurement task on the minute ventilation approach. We take inspiration from the area of network pruning to improve the efficiency of the model we apply \cite{subramaniam2020n2nskip}.

\section{Datasets}
Since there were no available datasets for our purpose of measuring minute validation remotely, we decided to create our own dataset. We chose 103 subjects with and without a history of breathing disorders consisting of 53 women and 50 men. All the protocols and procedures applied on the human subjects were applied with consent and approval of each subject. Each subject was made aware of the purpose of the study and willingly signed a written assent accepting to the same.

\section{Method}

\begin{figure}
\includegraphics[width=\textwidth]{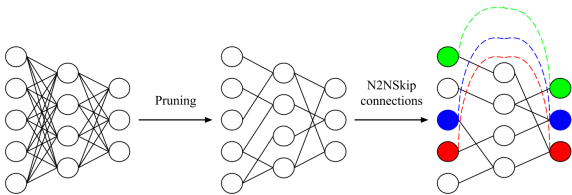}
\caption{Pattern of adding the skip connections to VGG-16 to improve the inference for minute ventilation with a more light weight network. } 
\label{fig:n2nskip}
\end{figure}

We propose two approaches to effectively measure minute ventilation outlined below:

\begin{itemize}
    \item NeuralNetA: The minute ventilation signals of the subject are obtained using the spirometer. However, this data is noisy since it contains considerable illumination and motion artifacts. Due to this, we propose an end to end solution to remove and clean the signal for further processing. We use an unpruned network similar to the structure of VGG16, which has over 13M parameters. The VGG network which we use is already pretrained on a private dataset containing breathing patters of the particular subject for whom we measure minute ventilation. We obtain over 400 such temporal patterns and segments for each subject. Our dataset consists of 103 subjects.  \\
    \item NeuralNetB: For our second approach NeuralNetB, we use a convolutional neural network similar to the structure of VGG-16. Since the vanilla VGG-16 has over 138M parameters which can be excessive for the task at hand, we opt to go for a one shot pruning strategy to reduce the number of parameters in the network and increase its efficiency. We choose to incorporate sparse convolutional skip connections to improve its overall connectivity, as shown in \cite{subramaniam2020n2nskip}. The pattern of adding these connections are shown in Fig. ~\ref{fig:n2nskip}. 
\end{itemize}

From Fig. \ref{fig:minute} and Fig. \ref{fig:minute2} we can see that the range of the RMSE is considerably lower in case of NeuralNetB (pruned network) as compared to NeuralNetA and the RMSE method. Additionally, the NeuralNetB has a much better correlation with the reference than the NeuralNetA model, for the estimation of the heart rate variability and minute ventilation.Furthermore, the discrepancy in the area of identity and the error are much more critical when the intensity of the illumination disturbances are larger.

\section{Experimental Results and Analysis}

\begin{figure}
\includegraphics[width=\textwidth]{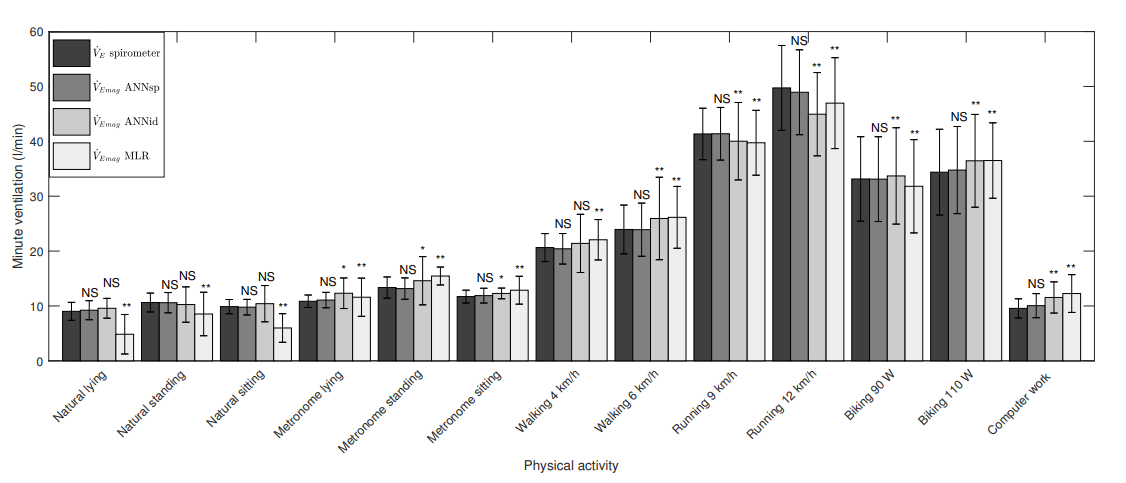}
\caption{Ground Truth values of minute ventilation based on estimation from a standard spirometer. NS represents a difference which is not statistically different.} 
\label{fig:minute2}
\end{figure}

\begin{figure}
\includegraphics[width=\textwidth]{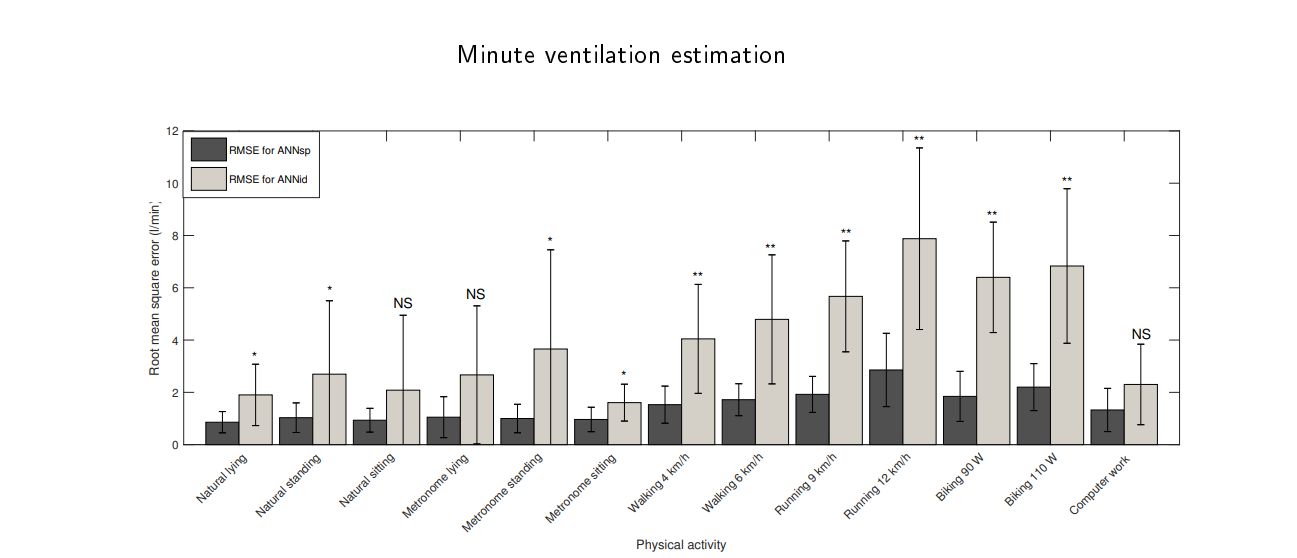}
\caption{The performance of our approach for both the approaches: NeuralNetA and NeuralNetB. NS represents a difference which is not statistically different.} 
\label{fig:minute}
\end{figure}

For our second approach NeuralNetB, we use a convolutional neural network similar to the structure of VGG-16. Since the vanilla VGG-16 has over 138M parameters which can be excessive for the task at hand, we opt to go for a one shot pruning strategy to reduce the number of parameters in the network and increase its efficiency. We choose to incorporate sparse convolutional skip connections to improve its overall connectivity, as shown in \cite{subramaniam2020n2nskip}. The pattern of adding these connections are shown in Fig. ~\ref{fig:n2nskip}. The pruned network is pretrained on CIFAR10 to ensure that we are able to transfer the learned parameters from the image classification task. 
We also compare the effectiveness of our approach against standard statistical measures such as root mean square error and find that our approach obtains superior performance in comparison to the aforementioned methods. 
From Fig. \ref{fig:minute} and Fig. \ref{fig:minute2} we can see that the range of the RMSE is considerably lower in case of NeuralNetB (pruned network) as compared to NeuralNetA and the RMSE method. Additionally, the NeuralNetB has a much better correlation with the reference than the NeuralNetA model, for the estimation of the heart rate variability and minute ventilation.Furthermore, the discrepancy in the area of identity and the error are much more critical when the intensity of the illumination disturbances are larger.

From Fig. \ref{fig:minute} and Fig. \ref{fig:minute2} we can see that the range of the RMSE is considerably lower in case of NeuralNetB (pruned network) as compared to NeuralNetA and the RMSE method. Additionally, the NeuralNetB has a much better correlation with the reference than the NeuralNetA model, for the estimation of the heart rate variability and minute ventilation.Furthermore, the discrepancy in the area of identity and the error are much more critical when the intensity of the illumination disturbances are larger.

\begin{figure}[t]
\includegraphics[width=\textwidth]{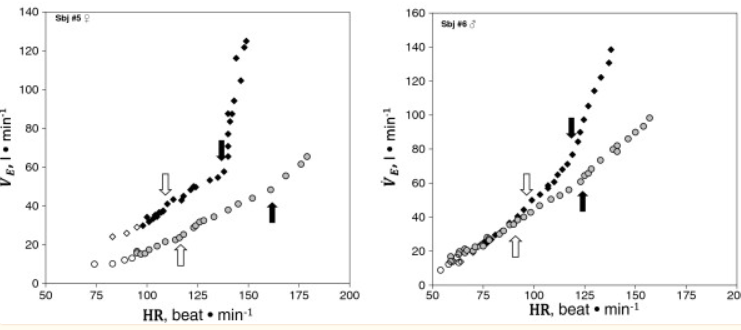}
\caption{The link between heart rate variability and minute ventilation for subjects between the age of 12 and 20 years of age, for carefree settings (left) and under duress (right).} 
\label{fig:ventilationGraph}
\end{figure}

\section{Conclusion}
We formulated a deep learning based approach to measure remote ventilation on a private dataset. The dataset
will be made public upon acceptance of this work. We used two versions
of a deep neural network to estimate the minute ventilation from data
streams obtained through wearable heart rate and respiratory devices.
We demonstrated that the simple design of our pipeline - which includes
lightweight deep neural networks - can be easily incorporate into real
time health monitoring systems. We applied two different kinds of approaches to measure the heart rate - NeuralNetA and NeuralNetB. From our experimental results, we found that NeuralNetB has a much
better correlation with the reference than the NeuralNetA model, for the estimation of the heart rate variability and minute ventilation. Furthermore, the discrepancy in the area of identity and the error are much more critical when the intensity of the illumination disturbances are larger.

\end{document}